\documentclass[sigconf]{acmart}

\renewcommand\footnotetextcopyrightpermission[1]{} 
\usepackage{multirow}
\usepackage{tabularx, colortbl}
\usepackage{balance}
\usepackage{pbox}
\usepackage{url}

\usepackage{hyperref}

\newcolumntype{C}[1]{>{\centering\arraybackslash}p{#1}}

\begin{document}

\setcopyright{none}


\author{Tommi Gr\"{o}ndahl}
\email{tommi.grondahl@aalto.fi}
\affiliation{Aalto University}

\author{Luca Pajola}
\email{luca.pajola@aalto.fi}
\affiliation{Aalto University}

\author{Mika Juuti}
\email{mika.juuti@aalto.fi}
\affiliation{Aalto University}

\author{Mauro Conti}
\email{conti@math.unipd.it}
\affiliation{University of Padua}

\author{N. Asokan}
\email{asokan@acm.org}
\affiliation{Aalto University}

\title[All You Need is ``Love'': Evading Hate Speech Detection]{All You Need is ``Love'': Evading Hate Speech Detection}

\begin{abstract}
With the spread of social networks and their unfortunate use for hate speech, automatic detection of the latter has become a pressing problem. In this paper, we reproduce seven state-of-the-art hate speech detection models from prior work, and show that they perform well only when tested on the same type of data they were trained on.
Based on these results, we argue that for successful hate speech detection, model architecture is less important than the type of data and labeling criteria.
We further show that all proposed detection techniques are brittle against adversaries who 
can (automatically) insert typos, change word boundaries or add innocuous words
to the original hate speech. A combination of these methods is also effective against Google Perspective -- a cutting-edge solution from industry.
Our experiments demonstrate that \textit{adversarial training} does not completely mitigate the attacks,
and using character-level features makes the models systematically more attack-resistant than using word-level features.
\end{abstract}

\maketitle

\section{Introduction}
\label{sec:Introduction}


Social networking has changed the way people communicate online.
While the ability of ordinary people to reach thousands of others instantenously undoubtedly has positive effects, downsides like polarization via echo chambers~\cite{Kumar:Shah2018} have become apparent. This inter-connectedness of people allows malicious entities to influence opinions by posting hateful material, also known as \emph{hate speech}.

Hate speech is not a universal concept. While laws targeting speech seen as harmful have existed throughout human civilization, the specific term was originally coined in the US in 1989 to address problems of ``harmful racist speech'' that was nonetheless protected in the US \cite{Brown2017}. In 1997, the European Union defined ``hate speech'' as texts that ``spread, incite, promote or justify racial hatred, xenophobia, antisemitism or other forms of hatred based on intolerance.''\footnote{\url{https://www.echr.coe.int/Documents/FS_Hate_speech_ENG.pdf}} Hate speech can be separated from merely offensive or shocking content \cite{Davidsonetal2017}, although this distinction is non-trivial. In this paper, we denote non-hateful speech as ``ordinary speech.''





Typically, hate speech detection is cast as a \emph{classification problem}. Standard machine learning algorithms are used to derive a discriminative function that can separate hate speech from ordinary speech. 
Although several hate speech detection mechanisms have been reported in the research literature~\cite{Schmidt:Wiegand2017}, to the best of our knowledge, there has so far been no systematic \textit{empirical} evaluation comparing actual implementations of proposed models and datasets.

We study five recent model architectures presented in four papers. One architecture \cite{Zhangetal2017} is trained separately on three different datasets, giving us seven models in total. Six of these distinguish between two classes \cite{Wulczynetal2017, Badjatiya2017, Zhangetal2017}. One classifies among three, distinguishing between offensive and non-offensive ordinary speech \cite{Davidsonetal2017}. Two models are \emph{character-based}~\cite{Wulczynetal2017} (using character n-grams as features) while the rest are \emph{word-based} (using word n-grams or embeddings).
In the original papers, each of the two-class models was evaluated using a particular dataset. We show that none of the pre-trained models perform well when tested with any other dataset. This suggests that the features indicative of hate speech are not consistent across different datasets. However, we also show that all models perform equally well if they are retrained with the training set from another dataset and tested using the test set from the same dataset. This suggests that hate speech detection is largely independent of model architecture.
We also tested each two-class model on offensive ordinary speech~\cite{Davidsonetal2017}, and observe that they tend to classify it as hate speech.
This indicates that the models fail to distinguish between hate speech and offensive ordinary speech, making them susceptible to false positives.
%
We experimented with using a \textit{transfer learning} approach~\cite{Howard:Ruder2018}. Using a pre-trained language model, we fine-tuned it for the classification tasks, and conducted the same experiments. We show that the results are comparable but do not exceed the baselines.


Prior work has only considered what can be called \textit{naive adversaries}, who do not attempt to circumvent detection. We show that all the models are vulnerable to adversarial inputs.
%
%
There are many ways of attacking text-based detection models. 
A simple attack involves changing the input text so that a human reader will still get the intended meaning, while detection models misclassify the text.
We suggest three such alteration techniques, all of which are easily automated:
(i) word changes, (ii) word-boundary changes, and (iii) appending unrelated innocuous words. Implementing two varieties of each attack, we show that all detection models are vulnerable to them, although to different extents.

Combining two of our most effective attacks, we present a simple but powerful evasion method, which completely breaks all word-based models, and severely hinders the performance of character-based models.
In addition, this attack significantly degreades the performance of Google Perspective API,\footnote{\url{https://www.perspectiveapi.com/}} which assigns a ``toxicity'' score to input text.

We summarize our contributions as follows:

\begin{itemize}

  \item The first experimental comparative analysis of state-of-the-art hate speech detection models and datasets
(Section~\ref{sec:Replication}).

  \item Several attacks against effective against all models and possible mitigations (Sections~\ref{sec:Attacks}).
  
  \item A simple but effective evasion method that completely breaks all word-based classifiers, and significantly impacts character-based classifiers as well as Google Perspective (Section~\ref{sec:Results-love}).
  



  \item A review of the limitations in current methds, and desiderata for future developments (Section~\ref{sec:Related}, Section~\ref{sec:Discussion}).

  \end{itemize}

\section{Replication and model comparison}
\label{sec:Replication}

We begin by describing four recent papers \cite{Wulczynetal2017,Davidsonetal2017,Badjatiya2017,Zhangetal2017} on hate speech detection (Section \ref{sec:Replication-models-datasets}). We reproduce and systematically analyze the performance of seven models presented in these papers (Section \ref{sec:Replication:model-performance}).
The datasets focus on partially disjoint aspects of "hate":
e.g., hate speech based on religion or ethnicity \cite{Zhangetal2017} may not be highly similar to sexually connotated hate speech~\cite{Badjatiya2017}.
As all datasets except one are drawn from Twitter, they do not properly represent the range of online text. We believe future research should focus on collecting and properly evaluating datasets, especially outside Twitter.

We tested the performance of all pre-trained models against four datasets. We further re-trained the proposed models on each of the other three datasets, and compared the results.
All model architectures performed comparably well, when the training set and test set were taken from the same dataset. 
However, our results indicate that hate speech detection is highly context-dependent and transfers poorly across datasets.


%

\subsection{Models and datasets}
\label{sec:Replication-models-datasets}
In this section, we describe all replicated models and their respective datasets. The datasets, labelled W, T1, T1*, T2, and T3, are summarized in Table~\ref{tab:datasets}, and Table~\ref{tab:data-stats} provides additional statistics on sentence lengths. All datasets were divided between a training set used to train the model, and a test set used for measuring model performace.

Each paper proposes a different machine learning model for hate speech detection: two papers use feature extraction -based models \cite{Wulczynetal2017,Davidsonetal2017} and two use recurrent neural networks \cite{Badjatiya2017,Zhangetal2017}. 
All models lowercase content and remove punctuation.
We summarize the models and datasets in Table~\ref{tab:models}, and discuss them in the remainder of this section.

\begin{table}[h]
  \begin{center}
    \begin{tabular}{|c|c|c|c|}
    \hline
    
    \textbf{Dataset} & \textbf{Domain} & \textbf{Classes (size)} & \textbf{Source} \\ \hline

    \multirow{2}{*}{W} & \multirow{2}{*}{Wikipedia} & personal attacks ($13590$) & \multirow{2}{*}{\cite{Wulczynetal2017}} \\ && ordinary ($102274$) & \\ \hline
    
    \multirow{3}{*}{T1} & \multirow{3}{*}{Twitter} & hate speech ($1430$) & \multirow{3}{*}{\cite{Davidsonetal2017}} \\ && offensive ($19190$) & \\ && ordinary ($4163$) & \\ \hline
    
    \multirow{2}{*}{T1*} & \multirow{2}{*}{Twitter} & hateful ($1430$) & \multirow{2}{*}{\cite{Davidsonetal2017}} \\ && offensive $\cup$ ordinary ($23353$) & \\ \hline

    \multirow{2}{*}{T2} & \multirow{2}{*}{Twitter} & racist $\cup$ sexist ($5013$) & \multirow{2}{*}{\cite{Waseemetal2016}} \\ && ordinary ($10796$) & \\ \hline
    
    \multirow{2}{*}{T3} & \multirow{2}{*}{Twitter} & hateful/racist ($414$) & \multirow{2}{*}{\cite{Zhangetal2017}} \\ && ordinary ($2021$) & \\ \hline
    
    \end{tabular}
    \caption{Datasets used in our replication. Union ($\cup$) denotes the conflation of class elements.}
    \label{tab:datasets}
  \end{center}
\end{table}

\begin{table}[h]
  \begin{center}
    \begin{tabular}{|c|c|c|c|c|c|c|c|}
    \hline
    
    \textbf{Dataset} & \textbf{Mean} & \textbf{Std.} & \textbf{Min.} & \textbf{Max.} & $\mathbf{25 \%}$ & $\mathbf{50 \%}$ & $\mathbf{75 \%}$ \\ \hline
    W & $84$ & $161$ & $1$ & $9949$ & $20$ & $42$ & $89$ \\ \hline
    T1, T1* & $20$ & $12$ & $1$ & $321$ & $11$ & $18$ & $27$ \\ \hline
    T2 & $20$ & $8$ & $1$ & $45$ & $14$ & $21$ & $26$ \\ \hline
    T3 & $20$ & $7$ & $1$ & $48$ & $16$ & $21$ & $25$ \\ \hline

    \end{tabular}
    \caption{Sentence lengths in the datasets.}
    \label{tab:data-stats}
  \end{center}
\end{table}

%
%

\begin{table}[h]
  \begin{center}
    \begin{tabular}{|c|c|c|c|}
    \hline
    
    \textbf{Model} & \textbf{Dataset(s)} & \textbf{Source} \\
    \hline
    
    LR char & W & \cite{Wulczynetal2017}\\ \hline
    MLP char & W & \cite{Wulczynetal2017} \\ \hline    
    LR word & T1 & \cite{Davidsonetal2017} \\ \hline
    CNN+GRU & T1*, T2, T3 & \cite{Zhangetal2017} \\ \hline
    LSTM & T2 & \cite{Badjatiya2017} \\ \hline
    
    \end{tabular}
    \caption{Replicated machine learning models.}
    \label{tab:models}
  \end{center}
\end{table}

\noindent\textbf{W \cite{Wulczynetal2017}}: 
As a part of Wikipedia's ``Detox'' project targeting personal attacks in Wikipedia's edit comments, Wulczyn et al. \cite{Wulczynetal2017} experimented with logistic regression (LR) and multilayer perceptron (MLP) models, using n-gram features on both the character- and word-level. Word n-gram sizes ranged from $1$ to $3$, and character n-gram sizes from $1$ to $5$. Labels were gathered via crowd-sourcing, each comment being labeled by $10$ evaluators. We denote this dataset as W.
Wulczyn et al. made their tests on both the one-hot encoded majority vote between the two classes (attack or non-attack), and empirical distributions based on different votes. The former gives a classification, whereas the latter may be interpreted as class probabilities.
On one-hot encoded labels, character-models performed better both in the original paper and our replication. We therefore use these models in our tests.

\noindent\textbf{{T1 \cite{Davidsonetal2017}}}: 
Davidson et al.~\cite{Davidsonetal2017} presented a dataset with three kinds of comments from Twitter: hate speech, offensive but non-hateful speech, and neither. This is, to our knowledge, the only dataset with such a distinction.
The hate speech data was collected by searching for tweets with known hate speech phrases,\footnote{The phrases were collected from \url{https://www.hatebase.org/}} and further labeling these tweets with a majority vote from three CrowdFlower workers each . We denote this dataset as T1. A vast majority of the dataset contains offensive speech (76\%), and only a small portion actual hate speech (5\%). 
Davidson et al. use a word-based logistic regression (LR) classifier ($1-3$-grams), which we replicated.

\noindent{\textbf{T2 \cite{Badjatiya2017}}}: 
In their paper, Wulczyn et al.~\cite{Wulczynetal2017} mention that future work should involve the use of deep neural networks (DNNs), which have proven useful in a number of text classification tasks \cite{Goldberg2016}. Currently, the dominant approaches to text classification use recurrent neural networks (RNNs), and in particular long short-term memory (LSTM) networks. These are applied for hate speech classification by Badjatiya et al.~\cite{Badjatiya2017}, who use them both alone and with gradient-boosting decision trees (GBDTs). For the latter, they first train the LSTM on the training data, and then use the average of the word embeddings learned by the LSTM as the input for the GBDT. They report a major increase in performance when compared with only using LSTMs. However, we were unable to replicate these results with the code they provide,\footnote{\url{https://github.com/pinkeshbadjatiya/twitter-hatespeech}} and received a better score using only the LSTM. Therefore, we use the LSTM alone instead of LSTM+GBDT. Like Badjatiya et al., we initialize the word embedding layer randomly.

As data, Badjatiya et al. use a corpus of $16 000$ tweets originally collected by Waseem et al. \cite{Waseemetal2016}. These contain three possible labels: ``racism'', ``sexism'', and ``neither''. In our experiments, we combined the ``racism'' and ``sexism'' classes into one, comprising general hateful material. We denote this dataset as T2. 

\noindent{\textbf{T1*, T3 \cite{Zhangetal2017}}}: 
In addition to LSTMs, convolutional neural networks (CNNs) have been popular in text classification research. Badjatiya et al. \cite{Badjatiya2017} experiment with them, as do Zhang et al. \cite{Zhangetal2017}. The latter add CNNs and RNNs together, by giving the output of a CNN to a gated recurrend unit (GRU) network. Word embeddings were initialized with Google word vectors trained on news data.\footnote{\url{https://github.com/mmihaltz/word2vec-GoogleNews-vectors}} 

Zhang et al. \cite{Zhangetal2017} use dataset T1 from Davidson et al.~\cite{Davidsonetal2017}, but combine offensive and ordinary speech into one category, as opposed to genuine hate speech. We denote this dataset as T1*. Zhang et al. further created their own small dataset of hate speech targeted toward refugees and muslims, which we denote as T3. Finally, in addition to T1* and T3, they experimented with T2. We replicate all three experiments, resulting in three distinct CNN+GRU models.

\subsection{Model performance}
\label{sec:Replication:model-performance}

In this section, we present replications of the original models (Section \ref{sec:Replication:replication}), and cross-apply all two-class models to all two-class datasets (Section \ref{sec:Replication-cross-application}).

\subsubsection{Replication and re-training}
\label{sec:Replication:replication}

In the original papers, the hyperparameters of each model have been optimized to the particular training sets used.
However, we additionally trained each model with \textit{every} dataset, and compared the results. As reported in Table~\ref{tab:Replication}, all models perform comparably on all four datasets.

\begin{table}[h]
  \begin{center}
    \begin{tabular}{|c|c|c|c|c|} \hline
    
     \multirow{2}{*}{\textbf{Model}} & \multicolumn{4}{c|}{\textbf{Dataset}} \\ \cline{2-5}
     & \textbf{W} & \textbf{T1*} & \textbf{T2} & \textbf{T3} \\ \hline
     LR char & $\mathbf{0.86}$ & $0.63$ & $0.82$ & $0.85$ \\ \hline
     MLP char & $\mathbf{0.86}$ & $0.63$ & $0.81$ & $0.85$ \\ \hline
     CNN+GRU & $0.87$ & $\mathbf{0.70}$ & $\mathbf{0.83}$ & $\mathbf{0.81}$ \\ \hline
     LSTM & $0.85$ & $0.64$ & $\mathbf{0.78}$ & $0.79$ \\ \hline    

    \end{tabular}
    \caption{F1-scores (macro-averaged across classes) of the two-class models trained and evaluated on each dataset (datasets used in original papers in bold).}
    
    \label{tab:Replication}
  \end{center}
\end{table}

The results indicate that the models are roughly equally effective when applied to different texts, provided that they were trained using the same kind of text. 

Inferior performance on T1* by all models can be explained by two factors. First, the dataset is highly imbalanced, with the ``hate'' class taking up only 5$\%$ of the training set. Second, this dataset is derived from Davidson et al.'s original three-class corpus that separates \textit{offensive speech} from hate speech. To constrain classification into only two categories, T1* assimilates the ``offensive'' and ``non-hate'' classes into one. Of these, the offensive class dominates, covering roughly 80$\%$ of the combined class in the training set. It may have more overlap with hate speech than non-offensive speech, making classification more challenging.

\subsubsection{Cross-application between datasets}
\label{sec:Replication-cross-application}

To estimate the adaptivity of models pre-trained with one dataset, we applied them to all test sets. We first trained each model with the training data used in the original paper presenting it, and then applied the resulting classifier to the test sets of all four datasets.

\begin{table}[h]
  \begin{center}
    \begin{tabular}{|c|c|c|c|c|c|} \hline
    
     \multirow{2}{*}{\textbf{Model, training dataset}} & \multicolumn{4}{c|}{\textbf{Dataset}} \\ \cline{2-5}
     & \textbf{W} & \textbf{T1*} & \textbf{T2} & \textbf{T3} \\ \hline
     LR char, W & $(0.86)$ & $0.37$ & $0.50$ & $0.24$ \\ \hline
     MLP char, W & $(0.86)$ & $0.38$ & $0.50$ & $0.25$ \\ \hline
     CNN+GRU, T1* & $0.11$ & $(0.70)$ & $0.48$ & $0.51$ \\ \hline
     CNN+GRU, T2 & $0.14$ & $0.28$ & $(0.83)$ & $0.44$ \\ \hline
     CNN+GRU, T3 & $0.13$ & $0.48$ & $0.50$ & $(0.81)$ \\ \hline
     LSTM, T2 & $0.23$ & $0.33$ & $(0.78)$ & $0.47$ \\ \hline

    \end{tabular}
    \caption{F1-scores (macro-averaged) of pre-trained two-class models applied to different test sets, with the original test set in parentheses.}
    
    \label{tab:Cross-application}
  \end{center}
\end{table}

As we can observe from the results reported in Table~\ref{tab:Cross-application}, none of the pre-trained models transfer well to any other dataset. These results indicate that linguistic hate speech indicators are not well retained across different datasets. This may be due to the lack of words shared between datasets, or differences between the relevant features of particular subcategories of hate speech.

\subsection{Offensive speech vs. hate speech}

An important distinction can be drawn between hate speech and speech that uses offensive vocabulary without being hateful. The latter is separated from hate speech in Davidson et al.'s dataset T1~\cite{Davidsonetal2017}.
Especially given the lack of any precise definition of hate speech in either legal or academic contexts, it is important to investigate the extent to which existing approaches are sensitive to the hateful-offensive distinction.


A particular concern is the extent to which two-class models assign offensive but non-hateful text to the ``hate'' class. This can be treated as a false positive, assuming the hateful-offensive distinction is appropriate. Of course, the assumption is not always clear: for instance, Google Perspective\footnote{\url{https://www.perspectiveapi.com/}} offers to detect ``toxic'' comments; but whether ``toxicity'' should include all offensive text remains a subjective assesment.

To estimate the performance of the models on offensive but non-hateful text, we applied the ``offensive'' datapoints from T1's test set to all two-class models, except CNN+GRU trained on T1*, which is built from T1 and hence not independent of the task.
As seen in Table~\ref{tab:Offensive}, CNN+GRU trained on T3 was the only model to succeed in this task. All other models performed on a random or below-random level.

\begin{table}[h]
  \begin{center}
    \begin{tabular}{|c|c|} \hline
    
     \textbf{Model, training dataset} & \textbf{Assignment to non-hate} \\ \hline
     LR char, W \cite{Wulczynetal2017} & $0.40$ \\ \hline
     MLP char, W \cite{Wulczynetal2017} & $0.36$ \\ \hline
     CNN+GRU, T2 \cite{Zhangetal2017} & $0.23$ \\ \hline
     CNN+GRU, T3 \cite{Zhangetal2017} & $0.89$ \\ \hline
     LSTM, T2 \cite{Badjatiya2017} & $0.36$ \\ \hline

    \end{tabular}
    \caption{Performance of two-class models on offensive but non-hateful speech.}
       
    \label{tab:Offensive}
  \end{center}
\end{table}


Additional experimentation revealed that the seeming success of CNN+GRU trained on T3 was due to the prevalence of unknown words in the ``offensive'' test set, mapped to the unknown token ($<$\textit{unk}$>$). This token was associated with the non-hateful class in the model, and on average over $40\%$ of words per sentence in this test set were mapped to it. Hence, the performance simply reflected the model's small vocabulary.

The results are suggestive of the problematic and subjective nature of what should be considered ``hateful'' in particular contexts. As the labels have been manually gathered via crowd-sourcing, there is no guarantee that labels from different people were consistent. Within datasets, this problem has can be addressed to some extent by majority voting, but there is no method of guaranteeing agreement between datasets. It seems that the labeling in the two-class models represents offensiveness more than hatefulness as such.

Manual experimentation suggests similar conclusions concerning Google Perspective. Some examples of Perspective's scores on non-hateful sentences appended with a common English curse-word (marked with ``F'' here, but in original form in the actual experiment) are presented in Table~\ref{tab:Curse}.

    
     
     
       

\begin{table}[h]
  \begin{center}
    \begin{tabular}{|l|c|} \hline    
     \textbf{Sentence} $\rightarrow$ \textbf{Modified sentence} & \textbf{Old} $\rightarrow$ \textbf{New score} \\ \hline
     You are great $\rightarrow$ You are F great & $0.03 \rightarrow 0.82$ \\
     I love you $\rightarrow$ I F love you & $0.02 \rightarrow 0.77$ \\ 
     I am so tired $\rightarrow$ I am F tired & $0.06 \rightarrow 0.85$ \\ 
     Oh damn! $\rightarrow$  Oh F! & $0.64 \rightarrow 0.96$ \\
     Food is amazing $\rightarrow$ Food is F amazing & $0.02 \rightarrow 0.68$ \\ \hline     
    \end{tabular}
    \caption{Google Perspective ``toxicity'' scores on non-hateful sentences with and without a curse word.}
       
    \label{tab:Curse}
  \end{center}
\end{table}

None of the examples in Table~\ref{tab:Curse} are hateful, but the curse-laden variants are all considered ``toxic'' with a high likelihood.
Clearly ``toxicity'', as Perspective currently classifies it, is not assimilable to hate speech in any substantive (or legal) sense.

\subsection{Transfer learning}

The replicated models use both classical machine learning methods, and more recent deep neural networks (DNNs). Recently, \textit{transfer learning} has been argued to improve performance text classification \cite{Howard:Ruder2018}. This approach to deep learning is based on fine-tuning a pre-trained model for a particular task, instead of training the entire model from scratch. A natural question to ask is whether performance on our datasets could be improved by using transfer learning.

We implemented a particular transfer learning method called \textit{ULMFiT} \cite{Howard:Ruder2018}, based on the code the authors provide.\footnote{\url{https://github.com/fastai/fastai}} The model uses a language-model pre-trained on a large dataset from Wikipedia \cite{Merityetal2017}, fine-tunes it for a particular dataset, and then fine-tunes a classifier on top of the language-model. We refer the reader to the original paper for technical details \cite{Howard:Ruder2018}.

Our preliminary results indicate that ULMFiT is unable to reach the baselines from the replicated models. On the Twitter datasets T1--T3, ULMFiT results ($0.62$, $0.75$, $0.80$) remained below all respective baselines ($0.66$, $0.84$, $0.86$). However, more systematic evaluation of transfer learning is needed in future research, and we are currently working on this question.

\subsection{Summary of replication and cross-application}

All four two-class models perform highly similarily when trained on each of the four datasets. In particular, neither the features used (characters vs. words) or model complexity influenced the test score in any significant way. This suggests that the features learned by all models (including LSTM and GRU models) are similar to relatively simple n-grams, as opposed to involving more complex relations between features.
\section{Attacks}
\label{sec:Attacks}


\noindent{\textbf{Adversary model}} The goal of the adversary we consider in our work is to fool a detection model into classifying hate speech input as ordinary speech. We assume that the adversary has complete control of the input and can modify it to evade detection while retaining the semantic content of the original hate speech. We do not assume that the adversary has whitebox access to the model parameters. The adversary model is relaxed in one attack, where some knowledge of the training set is required.

Classifiers that rely on surface-level features can be attacked with malformed input, constituting \textit{evasion attacks} \cite{Marpaungetal2012}.
In this section we describe six evasion attacks against the seven classifiers we replicated. All attacks are based around altering input text with easily implementable and automatic methods. We categorize the attacks into three types, and experiment with two alternatives from each:

\begin{itemize}

\item \textbf{Word changes}
\subitem{1.} Inserting typos
\subitem{2.} Leetspeak \cite{Pereaetal2008}

\item \textbf{Word-boundary changes}
\subitem{1.} Inserting whitespace
\subitem{2.} Removing whitespace

\item \textbf{Word appending}
\subitem{1.} Appending common words
\subitem{2.} Appending non-hateful words

\end{itemize}

We differentiate the attack types, because they target distinct (albeit sometimes overlapping) aspects of the classifier. Word changes are done to change the identities of words. For word-level models, the new identities are likely to become ``unknown'' (i.e. non-identified tokens denoted by $<$\textit{unk}$>$), while retaining the readability and semantic content of the original text to a maximal extent from a human reader's perspective. Word boundary changes, on the other hand, not only alter word identities but also the sentence structure from a superficial perspective. Finally, word appending makes no alterations to the original text, but only adds unrelated material to confuse classification.



There are theoretical reasons to believe certain attacks to be more effective against some classifiers than others. In particular, as word tokenization plays no role in character-models, it is clear that these are less susceptible to word-boundary changes than word-models.
Furthermore, character-models are expected to be more resistant to word changes, as many character n-grams are still retained after the transformations are applied. In contrast, all word-based models are expected to be vulnerable to attacks that alter word identities (i.e. both word changes and word-boundary changes).
Word appending attacks, in contrast, have no \textit{a priori} reason to work better against character- than word-based models, or vice versa.

We applied the attacks to all seven classifiers, by transforming the hate class samples from the respective test sets.

\subsection{Word changes}

Word change attacks introduce misspellings or alternative spellings into words. They may make words entirely unrecognizeable to word-based models,  
and change the distribution of characters in character-based classification.


In 2017, Hosseini et al. \cite{Hosseinietal2017} showed that Google's ``toxicity'' indicator Perspective could be deceived by typos. However, Perspective has since been updated, and the examples the authors provided no longer succeed to the extent they did then. We review these results in Section \ref{sec:Results-love}.
Still, we consider the proposal to be worth further investigation with larger datasets and automatic typo-generation.

In addition to typos, we also consider a simplified variety of \textit{leetspeak}, which is a type of Internet slang replacing characters with numbers. Different variants may involve more alterations, but we only consider character-to-number alterations here. Leetspeak has been shown to be easily readable to humans \cite{Pereaetal2008}, but will be unrecognizable to word-models unless it is also present in the training set.

\subsubsection{Algorithms}


\noindent \textbf{Inserting typos:}
As an attack, typo generation has three desiderata: (i) reducing the detection likelihood of hate speech, (ii) avoiding correction by automatic spell checkers, and (iii) retaining the readability and meaning of the original text. If the second goal is not met, the defender can include a spell-checker as a pre-processing stage in the classification. Satisfying the third goal requires the word to remain recognizable, and not to be mapped to some other word in the reader's mental lexicon.

To make the attack succesful, word-level changes must not only fool the classifier, but be human-readable and retain the original interpretation to a maximal extent. This means that we cannot simply introduce typos via random changes to words. Instead, the change must have minimal impact on readability. We utilize the empirical finding from cognitive psychology that characters in the middle of words have a smaller effect on readability than the first and last characters \cite{Rayneretal2006}. To maximize understandability, we restricted the alterations to a single switch between two characters.

The algorithm switches the order of two characters in the word, excluding the first and final character. The probability of choosing particular characters is calculated by two factors: characters closer to the middle of the word are preferred, and characters that are close to each other are preferred.
First, a character is chosen between the second and second-to-last character of the word, based on a Gaussian distribution centering in the middle. Hence, only words with four of more characters are applicable. Next, a second character is chosen based on the combined effect of the first distribution and a second Gaussian distribution centered around the previously chosen word.
Two random non-edge characters are consequently chosen, favoring characters in the middle of the word. These characters are then switched to create the typo.


\noindent \textbf{Leetspeak:}
To transform original text into simple leetspeak, we introduce the following character replacements:

\begin{center}
(a: $4$) (e: $3$) (l: $1$) (o: $0$) (s: $5$).
\end{center}


The changes retain readability well, given the visual similarity between the original and leet characters \cite{Pereaetal2008}.

\subsubsection{Mitigation}

We tried two methods of improving classifier performance on test data modified via word changes: \emph{adversarial training}, and adding a \emph{spell checker} to test set pre-processing.

\noindent{\textbf{Adversarial training:}} We augmented the training set with stochastically transformed versions of the original training examples, for all classes (doubling the training set size). The purpose of adversarial training was to add transformed varieties of each word into the model's vocabulary, making it more likely for the model to associate them with the correct class.

The random nature of the typo-algorithm limits the scalability of the approach with long words, as the range of possible typos becomes larger. In contrast, given the deterministic nature of our leetspeak algorithm, we can expect adversarial training to work well against it.

\noindent{\textbf{Spell checking:}} We added a spell-checker to the pre-processing stage of the test set to find out how resistant our typo-introduction algorithm was to automatic correction. We used Python's \textit{Autocorrect} library for this.

\subsection{Word boundary changes}

The tokenization attack differs from word transformation by retaining word-internal characters, but introducing or removing characters that result in a word-based model separating between different tokens. We use space as the most evident choice.

\subsubsection{Algorithms}

We implemented two simple algorithms for introducing or removing whitespace.
We predict removal to be more effective against word-based models on theoretical grounds. Character-based models are likely not highly susceptible to either variant.

\noindent \textbf{Inserting whitespace:}
Word-based models rely on tokenizing the text into a sequence of words, based on characters treated as word-separators. Therefore, a simple way to make words unrecognizable is to change the tokenization by introducing additional separators between words. The effect of this attack on readability is small, but it results in most words becoming unrecognized by the classifier.

We used a simple approach of splitting each (content-)word into two by randomly selecting a character from the word and adding a space before it. In a word-based model a previously recognized word will turn into $<$\textit{unk}$>$ $<$\textit{unk}$>$.

\noindent \textbf{Removing whitespace:}
Conversely, removing all spaces leaves a single $<$\textit{unk}$>$ datapoint. Word-based models' performance will then depend entirely on how this single token is classified. Character-models, in contrast, will only lose the information related to the space token, which might deteriorate their performance, but likely not much.

This attack has a marked negative impact on surface-level readability, but still allows the reader to recover the original content. We take the adversary's main goal to be getting his/her message across to their target, even if this required some effort on other end. As English sentences are rarely reformulable into other grammatical sentences only by changing whitespace distribution, ambiguity does not arise and information loss is marginal.

\subsubsection{Mitigation}


\noindent{\textbf{Adversarial training:}} For the whitespace insertion attack, we appended the training set with randomly split versions of the original data to include word-parts into the model's vocabulary (thus doubling the training set). Given that an $n$ character word can be split in two in $n-1$ possible ways, going through all possible combinations of splits in a sentence quickly results in a combinatory explosion. Hence, the effect of adversarial training is expected to be limited on analytic grounds.

For completeness, we also conducted adversarial training via space removal, although this is close to useless on the grounds that it only adds whole comments as single words to the model vocabulary, and associates them with one class. As this method can only have an effect if it encounters the exact comment again, it cannot scale and hence its relevance is close to none in word-based models. Character-models, in contrast, can make use of it, but its inclusion is mostly redundant, as the datapoints are not changed much in comparison to the originals.

\noindent{\textbf{Removing spaces}} Another mitigation method is available only for character-based models, where spaces are removed from both training and test data at the pre-processing stage. This method will by necessity remove any effect of adding spaces, but will also remove potentially relevant information from the data, as one common character is removed entirely. The effectiveness of this mitigation method thus depends on the relevance of the space character for the classification.

\subsection{Word appending}

All text classification is based on the prevalence of class-indicating properties in the data. Therefore, adding material that is indicative of one class oven another makes it more likely for the classifier to assign the sample to that class. In many tasks this is appropriate, but hate speech detection constitutes an important exception. Hate speech and non-hate speech are not reversible: adding hateful material to ordinary speech \emph{will} turn the text into hate speech, whereas adding non-hateful material to hate speech \emph{may} not change its status. This vulnerability invites a possible attack, where non-hateful material is inserted to hate speech to change the classification.

For our attack model, we assume that the adversary is aware of the use of additional unrelated words to distract automatic classifiers. Assuming also that the beginning of the additional material is clear enough from the discourse, readability and semantic retainment are secured.

\subsubsection{Algorithms}

We generated a random number (between $10$ and $50$) or words at the end of each text in the ``hate'' class of the test set. The words were randomly chosen from two possible sources, yielding two varieties of the attack.

\noindent{\textbf{Appending common English words:}} Google has provided a list containing the most common words appearing in a large book corpus \cite{GoogleWord}. We used random words from the top $1000$ of these words, excluding stopwords. The rationale behind this attack is that it requires no knowledge of the training data on the part of the attacker.
Further, the common English words are likely to be contained in the training corpus of many different datasets.

\noindent{\textbf{Appending common "non-hate" words:}} Here, we assume the attacker knows, or correctly guesses, a list of words in the training data's non-hateful class. He then appends the text with randomly chosen common (content) words from this class. We are still not assuming white-box access to the model itself.

\section{Results}
\label{sec:Results}

We performed our six attacks on each of the seven model-dataset combinations replicated from prior work. This yields $42$ attacks in total. We used adversarial training to mitigate all attacks, and tried additional defence methods for the typo attack (spell-checking) and tokenization attacks (space removal in training; this method is only available for character-models).

Attack effectiveness varied betweeen models and datasets, but the performance of all seven hate speech classifiers was significantly decreased by most attacks.
Word-based models were most affected by tokenization changes, and character-based models by word appending. A significant difference between word- and character-based models was that the former were all completely broken by at least one attack, whereas the latter were never completely broken. The two character-models performed comparably across all attacks.

The training dataset had a major influence on attack resilience, as demonstrated by the differences between the CNN+GRU model trained with T2 and T3, respectively.
However, CNN+GRU trained on T2 was more resilient to all attacks than LSTM trained on T2, indicating that model choice also had an effect.


The results from all attacks and their mitigation are presented in Table \ref{tab:Attacks}. Since our attacks affect datapoints in the ``hate'' class, we only report the F1-scores from this class.

\begin{table*}[t]
\begin{center}
\begin{tabular}
{|l|c|C{0.5cm} C{0.5cm} C{0.5cm} | C{0.5cm} C{0.5cm} | C{0.5cm} C{0.5cm} C{0.5cm} | C{0.5cm} C{0.5cm} C{0.5cm} | C{0.5cm} C{0.5cm} | C{0.5cm} C{0.5cm}|}

  \hline
  \multirow{3}{*}{\textbf{Model, DS}} &
  \multirow{3}{*}{\textbf{Orig.}} &
  \multicolumn{5}{c|}{\textbf{Word changes}} &
  \multicolumn{6}{c|}{\textbf{Boundary changes}} &
  \multicolumn{4}{c|}{\textbf{Word appending}} \\

  \cline{3-17}  
  
  & & \multicolumn{3}{c|}{\textbf{Typos}} &
  \multicolumn{2}{c|}{\textbf{Leet}} &
  \multicolumn{3}{c|}{\textbf{Insert}} &
  \multicolumn{3}{c|}{\textbf{Remove}} &
  \multicolumn{2}{c|}{\textbf{Common}} &
  \multicolumn{2}{c|}{\textbf{Non-hate}} \\
  
  & & \textbf{A} & \textbf{AT} & \textbf{SC} &
  \textbf{A} & \textbf{AT} &
  \textbf{A} & \textbf{AT} & \textbf{RW} &
  \textbf{A} & \textbf{AT} & \textbf{RW} &
  \textbf{A} & \textbf{AT} &
  \textbf{A} & \textbf{AT} \\
  \hline
  
  LR char, W & $0.75$ & $0.60$ & $0.71$ & $0.68$ & $0.61$ & $0.74$ & $0.75$ & $0.71$ &
  $0.74$ & $0.54$ & $0.75$ & $0.74$ & $0.48$ & $0.68$ & $0.47$ & $0.67$ \\ \hline
  
  MLP char, W & $0.75$ & $0.55$ & $0.71$ & $0.68$ & $0.59$ & $0.73$ & $0.75$ & $0.71$ &
  $0.72$ & $0.56$ & $0.76$ & $0.72$ & $0.50$ & $0.72$ & $0.48$ & $0.67$ \\ \hline  
  
  CNN+GRU, T1* & $0.43$ & $0.31$ & $0.35$ & $0.36$ & $0.00$ & $0.33$ & $0.04$ & $0.34$ & $-$ & $0.00$ \cellcolor[gray]{.8} & $0.00$ \cellcolor[gray]{.8} & $-$ & $0.04$ & $0.38$ & $0.01$ & $0.27$ \\ \hline  
  
  CNN+GRU, T2 & $0.76$ & $0.27$ & $0.67$ & $0.68$ & $0.09$ & $0.77$ & $0.43$ & $0.66$ & $-$ & $0.00$ \cellcolor[gray]{.8} & $0.00$ \cellcolor[gray]{.8} & $-$ & $0.64$ & $0.75$ & $0.50$ & $0.74$ \\ \hline
  
  CNN+GRU, T3 & $0.69$ & $0.23$ & $0.61$ & $0.43$ & $0.03$ & $0.76$ & $0.08$ & $0.63$ &  $-$ & $0.00$ \cellcolor[gray]{.8} & $0.00$ \cellcolor[gray]{.8} & $-$ & $0.18$ & $0.70$ & $0.14$ & $0.64$ \\ \hline
  
  LSTM, T2 & $0.70$ & $0.40$ & $0.66$ & $0.67$ & $0.19$ & $0.71$ & $0.42$ & $0.64$ &
  $-$ & $0.00$ \cellcolor[gray]{.8} & $0.02$ \cellcolor[gray]{.8} & $-$ & $0.27$ & $0.68$ & $0.15$ & $0.69$ \\ \hline
  
  LR word, T1 & $0.50$ & $0.30$ & $0.42$ & $0.37$ & $0.04$ & $0.48$ & $0.18$ & $0.44$ &
  $-$ & $0.01$ \cellcolor[gray]{.8} & $0.02$ \cellcolor[gray]{.8} & $-$ & $0.48$ \cellcolor[gray]{.8} & $0.44$ \cellcolor[gray]{.8} & $0.45$ \cellcolor[gray]{.8} & $0.30$ \cellcolor[gray]{.8} \\ \hline  
  
\end{tabular}

\caption{F1-scores on the ``hate'' class from attacks and mitigation. \\
Attack: A; Mitigations: \textbf{AT} = adversarial training, \textbf{SC} = spell-checker, \textbf{RW} = removing whitespace (character-models only) \\
Expected pattern is attack reducing score and mitigation restoring it; deviations highlighted and discussed in sections \ref{sec:Results-word-transformation}--\ref{sec:Results-word-appending}.}
\label{tab:Attacks}
\end{center}
\end{table*}

\subsection{Word changes}
\label{sec:Results-word-transformation}

Word-models were more susceptible to leetspeak than typos, whereas no clear difference can be found in character-models. In addition, word-models were much more vulnerable to both attacks than character-models.
Adversarial training had a major positive effect on performance against both attacks, but its effect was larger on the leetspeak attack.
This is unsurprising given the deterministic nature of the leetspeak algorithm. The determinacy also indicates that the leetspeak attack could easily be mitigated by a counter-algorithm transforming numbers into corresponding characters.

\subsection{Word boundary changes}
\label{sec:Results-tokenization}

Neither character-model was affected by the whitespace insertion attack, but the performance of both was markedly decreased by whitespace removal. We suggest this may be due to the fact that whitespace is involved in the beginnings and ends of words. Unlike adding whitespace, removing it abolishes all n-grams concerning word boundaries, which may be especially relevant for classification.

All word-models were completely broken by white space removal, and severely hindered by whitespace addition.
As expected, adversarial training had no impact on whitespace removal. Resistance to whitespace addition, in contrast, was improved, and reached levels close to the baseline, differing from it only $6-10 \%$.
Overall, removing whitespace was much more effective than its addition for all models, both as an attack and in avoiding adversarial training as mitigation.

Character-models peformed slightly worse when trained without spaces, but not much, the largest drop in F1-score being $3\%$. We conclude that removing spaces in pre-processing during both training and testing makes character-based models resistant to tokenization attacks with only a minor reduction in predictive power. No comparable mitigation exists for word-models, where word boundary removal will force the text to be tokenized as a single $<$\textit{unk}$>$.

\subsection{Word appending}
\label{sec:Results-word-appending}

Unlike other attacks, word appending affected character- and word-models comparably. Words from the non-hate class of the training set had a systematically larger impact than common English words, but the difference was very minor ($1-2\%$) on character-models. The largest difference was observed on the LSTM model, where non-hate words had almost twice the effect of common words ($0.27$ vs. $0.15$).

The only model not affected by either appending attack was the three-class word-based LR model from Davidson et al. \cite{Davidsonetal2017}, trained on T1. We attribute this result to the fact that the major non-hate class of this dataset was the ``offensive'' class. Common English words or words from the ``neither'' class rarely indicate offensive speech, making it unlikely for the hate speech to be classified as such. This data imbalance also likely explains the negative effect of adversarial training, which was not observed on any other model.

Adversarial training worked very well on all two-class models, reaching predictive power close to the baseline. The effect was the smallest with CNN+GRU trained on T1*, leaving $16\%$ behind the baseline with adversarial training. The dataset T1* is drawn from T1 by combining offensive and non-offensive ordinary speech into a single class. As offensive speech takes the overwhelming majority of T1, T1* is highly imbalanced. We therefore expect adversarial training to result in the appended words to associate with the ``non-hate'' class more readily than the ``hate'' class, which would account for its limited success in mitigation.

\subsection{Adding ``love''}
\label{sec:Results-love}

Finally, we present the results from our attack combining the two most powerful approaches we experimented with: whitespace removal and word appending.

Whitespace removal turns the sentence into a single $<$\textit{<unk>}$>$, making the classification entirely dependent on the model's prediction of this particular token. Models might behave differently with respect to it, and hence the effects of whitespace removal can be uncertain. This problem can be remedied by adding words strongly indicative of the non-hate class, effectively forcing the model to prefer it.

Furthermore, instead of using a long list of non-hateful words, we minimize the hindrance on readability by appending the text with only one word: ``love''. We choose this word because, intuitively, it is likely to negatively correlate with hate speech. Our results support this hypothesis.

\begin{table}[h]
  \begin{center}
    \begin{tabular}{|p{3cm}|C{1.5cm}|C{1.5cm}|C{1.5cm}|}
    \hline
    \textbf{Model (data, classes)} & \textbf{Original} & \textbf{``Love''} \\ \hline   
    LR char (W, 2) & $0.75$ & $0.52$ \\ \hline
    MLP char (W, 2) & $0.75$ & $0.55$ \\ \hline
    CNN+GRU (T1*, 2) & $0.36$ & $0.01$ \\ \hline
    CNN+GRU (T2, 2) & $0.79$ & $0.00$ \\ \hline
    CNN+GRU (T3, 2) & $0.78$ & $0.00$ \\ \hline
    LSTM (T2, 2) & $0.74$ & $0.00$ \\ \hline
    LR word (T1, 3) & $0.50$ & $0.00$ \\ \hline
    \end{tabular}
    \caption{F1-scores (macro) from the attack of removing word boundaries and adding ``love''.}
    \label{tab:Attack-love}
  \end{center}
\end{table}

As expected, the attack completely broke all word-models, and significantly hindered character-models (Table \ref{tab:Attack-love}). Based on the word appending results, the performance on character-models would decrease with more appended non-hateful words.
Since the original message can relatively easily be recovered, and the single unrelated word has a minimal effect on readability (assuming it to be separable from the rest by the reader), we consider this attack highly successful against state-of-the-art hate speech classification.

We additionally experimented with Google Perspective, using example sentences from Hosseini et al. \cite{Hosseinietal2017}, reproduced below.

\begin{itemize}
\item[\textbf{S1:}] Climate change is happening and it's not changing in our favor. If you think differently you're an idiot.
\item[\textbf{S2:}] They're stupid, it's getting warmer, we should enjoy it while it lasts
\item[\textbf{S3:}] They are liberal idiots who are uneducated
\item[\textbf{S4:}] idiots. backward thinking people. nationalists. not accepting facts. susceptible to lies.
\item[\textbf{S5:}]  They are stupid and ignorant with no class
\item[\textbf{S6:}] It's stupid and wrong
\item[\textbf{S7:}] If they voted for Hilary they are idiots
\item[\textbf{S8:}] Anyone who voted for Trump is a moron
\item[\textbf{S9:}] Screw you trump supporters
\end{itemize}

Hosseini et al. provide examples of introducing manual typos, punctuation and spaces in each sentence, and demonstrate its effects on Google Perspective. We applied the same modifications to the same example sentences, but received much higher toxicity scores. This indicates that Perspective has been updated, and now performs better on adversarial data. Nevertheless, we managed to reduce the toxicity score to below $0.40$ on all example sentences with our attack of removing word boundaries and adding ``love'' (Table \ref{tab:Attack-Perspective}). This number is Perspective's upper limit for the label ``unlikely to be considered toxic''.


\begin{table}[h]
  \begin{center}
    \begin{tabular}{|C{1.5cm}|C{1.5cm}|C{1.5cm}|C{1.5cm}|}
    \hline
    \multirow{2}{*}{\textbf{Sentence}} & \multirow{2}{*}{\textbf{Original}} & \multicolumn{2}{c|}{\textbf{Modified}} \\ \cline{3-4}
    && \textbf{\cite{Hosseinietal2017}} & \textbf{``Love''} \\ \hline
    S1 & $0.95$ & $0.94$ $(0.20)$ & $0.35$ \\ \hline
    S2 & $0.92$ & $0.38$ $(0.02)$ & $0.37$ \\ \hline
    S3 & $0.98$ & $0.79$ $(0.15)$ & $0.37$ \\ \hline
    S4 & $0.95$ & $0.90$ $(0.17)$ & $0.37$ \\ \hline
    S5 & $0.97$ & $0.53$ $(0.11)$ & $0.37$ \\ \hline
    S6 & $0.88$ & $0.82$ $(0.17)$ & $0.35$ \\ \hline
    S7 & $0.99$ & $0.70$ $(0.12)$ & $0.15$ \\ \hline
    S8 & $0.96$ & $0.64$ $(0.13)$ & $0.35$ \\ \hline
    S9 & $0.90$ & $0.78$ $(0.17)$ & $0.35$ \\ \hline
    \end{tabular}
    \caption{Google Perspective ``toxicity'' scores for S1--S9. \\
    Scores from manual modifications \cite{Hosseinietal2017} in third column; results reported in the original paper in parentheses.}
    \label{tab:Attack-Perspective}
  \end{center}
\end{table}

\section{Discussion}
\label{sec:Discussion}

We evaluated the performance of seven state-of-the art hate speech classifiers presented in prior work. We showed that these techniques work roughly equally with different datasets, provided that the training and testing are based on the same dataset. However, we identified three main deficiencies in the models: (i) lack of effective transferability across datasets, (ii) conflation of hate speech and offensive ordinary speech, and (iii) susceptibility to simple text modification attacks. The first two arise from the problematicity of the concept of ``hate speech'', which can differ across datasets, and may or may not include all offensive or inappropriate material depending on the context.

Our attacks were much more effective against word-models than character-models. Most effective was the very simple ``love'' attack, which managed to completely break all word-models, and severely hinder the performance of character-models. We further demonstrated the attack's ability to reduce Google Perspective's toxicity score to below the threshold of $0.40$ in all our example sentences.

In this section we present some consequences for future work that, we suggest, are implied by our findings.

\subsection{Transferability and false positives}

No two-class model performed well on other datasets, and all of them classify most offensive ordinary speech as hateful. Manual experimentation showed that Google Perspective functions similarily, as adding curse words to otherwise benign text drastically increases the toxicity score. These results are indicative of two related problems.

First, the standards of ground-truth labeling likely varied across different datasets. For example, catagories like ``sexism/racism'' (T2) might be appropriate for some comments which are not ``personal attacks'' (W), or vice versa. This problem may not be fatal to institutions that wish to target particular subtypes of hate speech, as long as appropriate labels are available for sufficient training data. Our cross-application results further indicate that, for classification performance, model type matters less than dataset. However, the problem is more serious for the task of more general hate speech detection, as undertaken in law enforcement or academic research.

Second, with the exception of Davidson et al. \cite{Davidsonetal2017}, the distinction between hate speech and more generally ``inappropriate'' material is typically not made clear.
Google Perspective does not distinguish between hate speech and offensive speech, characterizing their ``toxicity'' metric as the means to identify a ``rude, disrespectful or unreasonable comment that is likely to make you leave a discussion''.\footnote{\url{https://www.perspectiveapi.com}}
Hence, the problem is not only that offensive ordinary speech can constitute a false positive. Rather, it is not clear where the boundary between true and false positives should lie.

We can, however, assume that a datapoint constitutes a genuine false positive at least when the only reason it is allocated to the hate class is because it contains curse words not used to offend any particular person or group. For future work, we therefore recommend experimenting with methods that discard curse words at least on the unigram level.

\subsection{Evasion attacks}

Our results show that character-models are much more resistant to simple text-transformation attacks against hate speech classification. While this is not theoretically surprising, to our knowledge it has not been taken into account in prior work on the topic. In particular, space-removal during training has only a minor negative effect on performance, mitigates all tokenization attacks, and is only available for character-models. Further, while we received the best results overall on CNN+GRU, the simple character-models fared reasonably well in comparison to word-based DNNs. We conclude that using character-models instead of word-models is the most effective protection against our attacks.

Nevertheless, all models were susceptible to the word appending attack. This was expected, since the other attacks are built around the transformation of word identities, but the appending attack has a different basis. It takes advantage of a fundamental vulnerability of all classification systems: they make their decision based on \textit{prevalence} instead of \textit{presence}. The status of some text as hateful is a matter of it containing some hateful material; whether this material constitutes the majority of the text is irrelevant. Classification, in contrast, looks at the entire sentence and makes its decision based on which class is more represented on average.
In principle, any text classification system can be forced to make a particular prediction by simply adding enough material indicative of one class.

We can re-phrase the problem by referring to the distinction between classification and \textit{detection}, where the latter consist in finding some features of a relevant sort, regardless of the prevalence of other material in the datapoint. In particular, we suggest re-conceptualizing hate speech detection as \textit{anomaly detection}, where hate speech constitutes an anomalous variant of ordinary speech.

As our attacks target commonly used  model architectures in text classification, they are generalizable beyond the task of hate speech detection. Possible other attack scenarios involve e.g. fooling sentiment analysis, author anonymization \cite{Brennan:Greenstadt2009, Brennanetal2011}, or avoiding content-based text classification to escape censorship.

On the other hand, our attacks only concern textual content, and hence do not hinder hate speech detection methods based around meta-features concerning user behavior \cite{Schmidt:Wiegand2017}. Given both the simplicity and effectiveness of our attacks, focusing on meta-level approaches instead of text classification can be a useful direction for future research.

\section{Ethical considerations}
\label{sec:Ethical-considerations}

For our replication and cross-application, we only used freely available online datasets and models. We collected no data ourselves, and none of our tests involved human subjects.
Since our original code constitutes a series of attacks, we do not provide it open source.
However, we will make it available for \textit{bona fide} researchers to facilitate reproducibility.
\section{Related work}
\label{sec:Related}


In their survey on hate speech detection, Schmidt and Wiegand \cite{Schmidt:Wiegand2017} categorize the features used in prior research into eight categories:

\begin{itemize}
\item[(i)] simple surface features
\item[(ii)] word generalization
\item[(iii)] sentiment analysis
\item[(iv)] lexical resources
\item[(v)] linguistic features
\item[(vi)] knowledge-based features
\item[(vii)] meta-information
\item[(viii)] multimodal information
\end{itemize}

Focusing only on linguistic features, we disregard (vii)--(viii).

Of simple surface features, character n-grams have been argued to perform better than word n-grams, since they can detect similarities between different spelling variants \cite{Mehdad:Tetreault}. These results are in line with ours. Word generalization has traditionally involved word clustering, i.e. assimilating similar words \cite{Warner:Hirschberg2012}, and more recently word embeddings. However, the superiority of embeddings over n-grams is not empirically attested, as both character and word n-grams have performed better when compared with embeddings in hate speech classification \cite{Nobataetal2016}.

DNNs typically include a word embedding layer in the beginning of the network, which is also true of the models we experimented with (LSTM, CNN+GRU). Prior to training, the embedding layer can be initialized randomly, or initialized by pre-trained embeddings like word2vec \cite{Word2Vec2013} or GloVe \cite{Pennington2014} Of the models we used, the LSTM \cite{Badjatiya2017} embeddings were randomly initialized, whereas the CNN+GRU \cite{Zhangetal2017} embeddings were initialized with Google embeddings trained on a news corpus.\footnote{\url{https://github.com/mmihaltz/word2vec-GoogleNews-vectors}}

Sentiment analysis can be incorporated to the process either as a prior filter \cite{Gitarietal2015}, or as a feature used directly for hate speech classification. Of the models we experimented with, the three-class LR-model of Davidson et al. \cite{Davidsonetal2017} includes sentiment as a textual feature. One important domain of future work involves applying our attacks on state-of-the-art sentiment classifiers to see if they can be broken to the same extent with simple text transformation methods.

The remaining (linguistic) feature-types (iv)--(vi) consist of more traditional, often hand-crafted, features and rules. Lists of hateful words are available online,\footnote{The most extensive of such lists are currently found in \url{https://www.hatebase.org/}.} and can be appended to other features in aid of hate speech detection. As stand-alone features, their performance is weak in comparison to n-grams \cite{Nobataetal2016, Schmidt:Wiegand2017}.
Of linguistic features applied to hate speech classification, the most common have been part-of-speech tags and dependency relations \cite{Chenetal2012, Burnap:Williams2015, Nobataetal2016}. Knowledge-based approaches based on automatic reasoning can help in detecting particular patterns related to hate speech \cite{Dinakaretal2012}, but do not scale beyond those patterns.

A general trend within NLP in recent years has been a shift toward using DNNs as opposed to more traditional keyword- and rule-based methods, or traditional machine learning approaches building on simple sparse features like n-grams \cite{Goldberg2016}.
However, our attack results indicate that reconsidering some older methods could be useful, as they may be more resistant toward the word appending attack. In particular, keyword-based approaches are not vulnerable to the class asymmetry problem, as the mere presence of hate-indicative keywords is relevant, irrespective of the presence of other words.


Outside of hate speech detection, text obfuscation -based evasion attacks have been conducted to avoid \textit{spam detection}, especially Bayesian models \cite{Sternetal2004}. In particular, our word appending attack bears a close similarity to what have been called ``good word attacks'' on spam filters \cite{Lowd:Meek2005, Zhouetal2007}. Here, the spammer injects words that are treated as indicative of legitimate text by the filter, with the goal of flipping the model's class prediction. Despite these well-known attacks on spam detection, analogical cases for hate speech have so far been neglected in the literature. We hope to alleviate this problem.

\section{Conclusions and future work}

Our replication and cross-application results suggest that model architecture has no major impact on classifier performance. Additionally, the simplest model (LR-char) performed comparably to more complex models, indicating that the positive effect of using more complex architectures is only minor. Cross-application further demonstrated that model complexity did not help to improve scalability across datasets. Instead, the problem stems from the labels themselves, the grounds of which can differ between the datasets.

We therefore suggest that future work should focus on the datasets instead of the models. More work is needed to compare the linguistic features indicative of different kinds of hate speech (racism, sexism, personal attacks etc.), and the differences between hateful and merely offensive speech.

The effectiveness of our simple attacks is indicative of the vulnerability of proposed hate speech classifiers based on state-of-the-art machine learning approaches. Future work should take such attacks into consideration in addition to mere classification accuracy. In particular, we demonstrated the superiority of character-models against attacks, which provides a significant case in favor of using them in real-world applications.

The appending attack presents a fundamental problem with treating hate speech detection as classification. The classes are asymmetrical in that ordinary speech can be transformed into hate speech by adding hateful material, but hate speech should not be transformed into ordinary speech by adding benign material. This asymmetricity is not built into classification, but it should be a foundational principle of hate speech detection. We recommend focusing on this problem in future research, and seeking detection methods that are not based on mere classification.
One possibility is to reintroduce more traditional keyword-based approaches, where only hate-indicative words are sought, disregarding the presence or absence of other words.

Additionally, building on our adversarial training experiments, we suggest training data augmentation as a method to help classification remain more resistant against appending attacks. This is a well-known approach to making classifiers more resistant to noise. Adding benign text to hate speech datapoints helps the classifier find those aspects that are relevant for the text belonging to the hate class, and decreases the effect of irrelevant word-class correlations.

In summary, we make four recommendations for future work on hate speech detection:

\begin{itemize}
\item Focus should be on the datasets instead of the models, and more qualitative work is needed to understand different categories the fall under the umbrella of ``hate speech''.

\item Simple character-models are preferable to word-based models (including DNNs) with respect to resisting simple evasion methods based on text transformation.

\item Detection methods should not be vulnerable to the asymmetricity between the classes, which invites using methods that only target the presence of hate-indicative features and remain indifferent to other features.

\item Training data augmentation can reduce the effect of benign words on classification.

\end{itemize}


\balance
\bibliography{hs-refs}
\end{document}